\documentclass{article}

\usepackage{PRIMEarxiv}
\usepackage{float}
\usepackage[utf8]{inputenc} 
\usepackage[T1]{fontenc}    
\usepackage{amsmath}        
\usepackage{hyperref}       
\usepackage{url}            
\usepackage{booktabs}       
\usepackage{amsfonts}       
\usepackage{nicefrac}       
\usepackage{microtype}      
\usepackage{lipsum}
\usepackage{fancyhdr}       
\usepackage{graphicx}       
\graphicspath{{media/}}     

\title{TrackNetV5: Residual-Driven Spatio-Temporal Refinement and Motion Direction Decoupling for Fast Object Tracking
\thanks{\textit{\underline{Preprint Citation}}:
\textbf{Tang, H., Chen, Y., and Jiang, L. TrackNetV5: Residual-Driven Spatio-Temporal Refinement and Motion Direction Decoupling for Fast Object Tracking. Submitted to ArXiv (2025).}}
}

\author{
  Haonan Tang\thanks{Equal contribution} \\
  Wuhan University of Technology \\
  Shanghai Code Zero Sports Technology Co. \\
  \texttt{lancera.thn@gmail.com} \\
  \And
  Yanjun Chen\thanks{Equal contribution} \\
  Wuhan University of Technology \\
  \texttt{mschen250@gmail.com} \\
  \And
  Lezhi Jiang\thanks{Equal contribution} \\
  Wuhan University of Technology \\
  \texttt{jackjianglezhi@gmail.com} \\
  \And
  Qianfei Li\thanks{Corresponding author} \\
  Shanghai Code Zero Sports Technology Co. \\
  \texttt{qf.qianfei@gmail.com} \\
  \And
  Xinyu Guo\thanks{Corresponding author} \\
  Shanghai Code Zero Sports Technology Co. \\
  \texttt{xinyg1412@gmail.com}
}

\begin{document}
\maketitle

\begin{abstract}
The TrackNet series has established a strong baseline for fast-moving small object tracking in sports. However, existing iterations face significant limitations: V1–V3 struggle with occlusions due to a reliance on purely visual cues, while TrackNetV4, despite introducing motion inputs, suffers from \textit{directional ambiguity} as its absolute difference method discards motion polarity. To overcome these bottlenecks, we propose TrackNetV5, a robust architecture integrating two novel mechanisms. First, to recover lost directional priors, we introduce the Motion Direction Decoupling (MDD) module. Unlike V4, MDD decomposes temporal dynamics into signed \textit{polarity fields}, explicitly encoding both movement occurrence and trajectory direction. Second, we propose the Residual-Driven Spatio-Temporal Refinement (R-STR) head. Operating on a coarse-to-fine paradigm, this Transformer-based module leverages factorized spatio-temporal contexts to estimate a corrective residual, effectively recovering occluded targets. Extensive experiments on the TrackNetV2 dataset demonstrate that TrackNetV5 achieves a new state-of-the-art F1-score of \textbf{0.9859} and an accuracy of \textbf{0.9733}, significantly outperforming previous versions. Notably, this performance leap is achieved with a marginal 3.7\% increase in FLOPs compared to V4, maintaining real-time inference capabilities while delivering superior tracking precision.
\end{abstract}


\section{Introduction}

High-speed small object tracking, particularly for tennis balls, is a cornerstone of modern sports analytics. It facilitates critical applications such as electronic line calling, tactical analysis, and broadcast visualization. However, this task presents severe computer vision challenges: targets are subjected to extreme motion blur, drastic deformation, variable lighting, and frequent occlusion by athletes. Consequently, traditional visual tracking methods often struggle to maintain robust localization under such dynamic conditions.

With the advent of deep learning, the TrackNet series \cite{2019TrackNetv1,2020TrackNetV2,2023TrackNetV3p,2023TrackNetV3up,2024TrackNetV4} has emerged as the de facto benchmark in this domain. TrackNetV1 \cite{2019TrackNetv1} pioneered the heatmap-based Encoder-Decoder paradigm but suffered from limited recall. TrackNetV2 \cite{2020TrackNetV2} substantially improved throughput and accuracy via the Multiple-In-Multiple-Out (MIMO) architecture. More recently, TrackNetV3 \cite{2023TrackNetV3up} demonstrated that stacking heavy computational modules (e.g., Inception-style blocks) could achieve higher accuracy, but at the cost of significantly increased latency (approx. $3\times$ slower than V2), rendering it less viable for resource-constrained real-time applications.

To balance efficiency and accuracy, the state-of-the-art TrackNetV4 \cite{2024TrackNetV4} integrated motion priors based on absolute frame differences. While efficient, V4 and previous iterations still exhibit distinct structural limitations that hinder further performance breakthroughs:

\begin{enumerate}
    \item \textbf{Directional Ambiguity:} TrackNetV4 extracts motion cues using absolute frame differences ($|I_t - I_{t-1}|$). While this captures movement occurrence, the absolute value operation inherently discards the \textit{sign} of intensity changes (polarity). Consequently, the network loses critical directional priors—specifically, the vector pointing from the "darkening" departure region to the "brightening" arrival region—which are essential for distinguishing the ball from background noise.
    
    \item \textbf{Underutilized Spatio-Temporal Context:} Existing MIMO architectures (V2–V4) generate heatmaps for consecutive frames (e.g., $t-1, t, t+1$) in a single pass. However, these predictions are generated as independent structural outputs without a mechanism to enforce temporal consistency. The latent correlation between consecutive heatmaps—where a high-confidence detection in one frame should support low-confidence detections in adjacent frames—remains unexploited.
\end{enumerate}

To address these limitations, we propose TrackNetV5, a novel architecture that integrates explicit directional modeling with coarse-to-fine refinement. Our contributions are twofold:

First, we introduce the Motion Direction Decoupling (MDD) module. Addressing the directional ambiguity in V4, MDD decomposes raw temporal dynamics into distinct positive and negative \textit{polarity fields}. This mechanism explicitly disentangles movement occurrence from movement direction, providing the encoder with robust directional cues without increasing model depth.

Second, we propose the Residual-Driven Spatio-Temporal Refinement (R-STR) head. Instead of treating the decoder's output as the final result, we view it as a preliminary "draft." The R-STR module leverages factorized spatio-temporal attention to estimate a residual correction map, effectively refining the draft by recovering occluded targets and suppressing motion artifacts based on temporal coherence.

Extensive experiments on the TrackNetV2 dataset demonstrate that TrackNetV5 achieves a new state-of-the-art performance with an F1-score of \textbf{0.9859} and an accuracy of \textbf{0.9733}, significantly outperforming all prior versions. Notably, this performance leap is achieved with a marginal 3.7\% increase in FLOPs compared to V4, maintaining the series' signature efficiency while delivering superior tracking precision.

\section{Related Work}

\subsection{Small Object Tracking in Sports}

The \textbf{TrackNet series} \cite{2019TrackNetv1,2020TrackNetV2,2023TrackNetV3p,2023TrackNetV3up,2024TrackNetV4} has established the dominant paradigm for high-speed ball tracking. TrackNetV1 \cite{2019TrackNetv1} introduced the heatmap regression framework, treating tracking as a specific localization task. TrackNetV2 \cite{2020TrackNetV2} advanced this by employing a Multiple-In-Multiple-Out (MIMO) architecture to process consecutive frames, significantly improving throughput. Subsequent variants, such as TrackNetV3 \cite{2023TrackNetV3up}, focused on maximizing accuracy by stacking heavyweight modules (e.g., Inception blocks), but at the cost of real-time performance.

Most recently, TrackNetV4 \cite{2024TrackNetV4} integrated motion priors via absolute frame differences, successfully highlighting regions of movement. However, as discussed in Sec. 1, its reliance on absolute magnitudes inherently discards \textit{directional polarity}, limiting the model's ability to distinguish object trajectories from background noise. In contrast, our TrackNetV5 introduces a polarity-aware motion decoupling mechanism to recover these lost directional cues while maintaining high efficiency.

\subsection{Motion Representation in Video Analysis}

Motion representation is critical for distinguishing moving targets from complex backgrounds. Existing approaches generally fall into two categories:

\textbf{Explicit Motion Modeling.} Methods like Optical Flow explicitly calculate pixel-level displacement vectors. Recent works, such as TOTNET \cite{2025TOTNet}, combine optical flow maps with 3D convolutions to extract robust spatio-temporal features. While accurate, optical flow computation incurs a prohibitive computational overhead, rendering it unsuitable for latency-sensitive applications.

\textbf{Implicit Motion Modeling.} Alternatively, architectures like I3D \cite{2018I3Dv3} or Video Transformers \cite{2021TimeSformer}\cite{2022VideoSwin} utilize 3D convolutions or self-attention to implicitly learn spatio-temporal dependencies. Although effective, these models typically require massive parameter counts and heavy computation (FLOPs).

To balance efficiency and accuracy, lightweight approaches like TrackNetV4 utilize frame differencing. While computationally inexpensive, standard differencing is a "lossy" operation that obscures motion direction. Our proposed \textbf{Motion Direction Decoupling (MDD)} module bridges this gap. It offers the directional clarity of optical flow-like representations but retains the computational efficiency of simple frame differencing.

\subsection{Spatio-Temporal Refinement}

Refinement strategies are widely used to correct initial predictions. In static keypoint detection, Cascaded Architectures (e.g., Convolutional Pose Machines \cite{2016CPM}) refine heatmaps stage-by-stage. However, these spatial-only methods operate on single frames, ignoring the temporal consistency inherent in video data.

In the video domain, Temporal Attention mechanisms (e.g., TimeSformer \cite{2021TimeSformer}) excel at capturing long-range dependencies. However, existing video Transformers are typically designed as heavyweight backbones rather than lightweight refinement heads. Furthermore, most refinement approaches employ a \textit{reconstruction} paradigm, forcing the network to regenerate the entire heatmap from scratch.

Our \textbf{Residual-Driven Spatio-Temporal Refinement (R-STR)} addresses these limitations. Unlike spatial cascades, it exploits temporal context from the MIMO output. Unlike heavy video backbones, it is a lightweight head designed for efficiency. Crucially, by adopting a \textit{residual learning} formulation, it simplifies the optimization landscape, focusing solely on correcting local errors rather than full-scale reconstruction.

\section{Methodology}

\subsection{Overall Architecture}

\begin{figure}[h]
  \centering
  \includegraphics[width=0.9\linewidth]{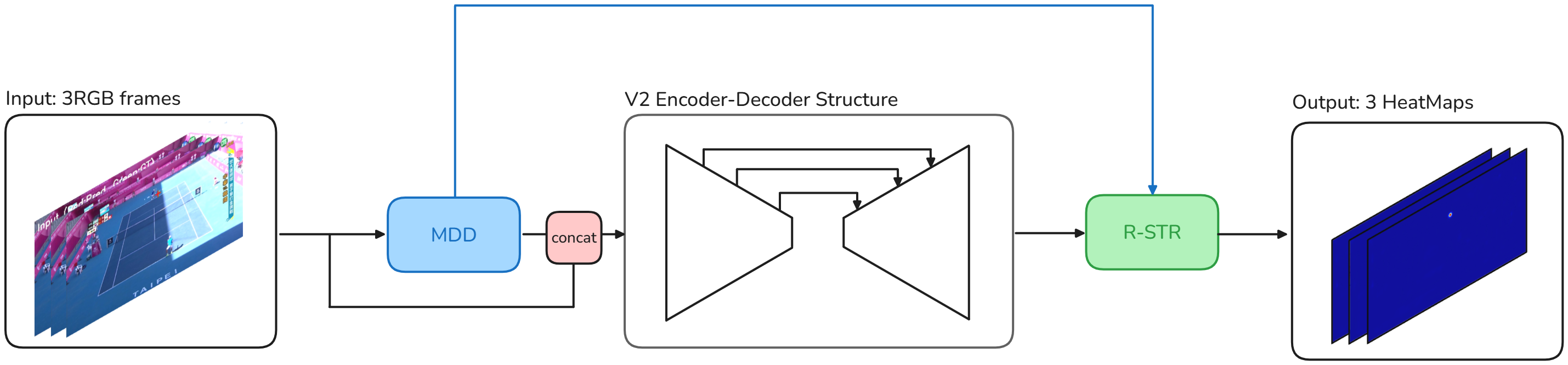} 
  \caption{Overall Architecture of TrackNetV5. }
  \label{fig:tracknetv5_overall} 
\end{figure}

The TrackNetV5 framework, illustrated in Figure \ref{fig:tracknetv5_overall}, is designed as a unified end-to-end pipeline that integrates explicit motion modeling with coarse-to-fine refinement. The architecture effectively extends the baseline V2 Encoder-Decoder backbone by incorporating the proposed MDD and R-STR modules at the logical input and output stages, respectively.

\begin{figure}[h]
  \centering
  \includegraphics[width=0.9\linewidth]{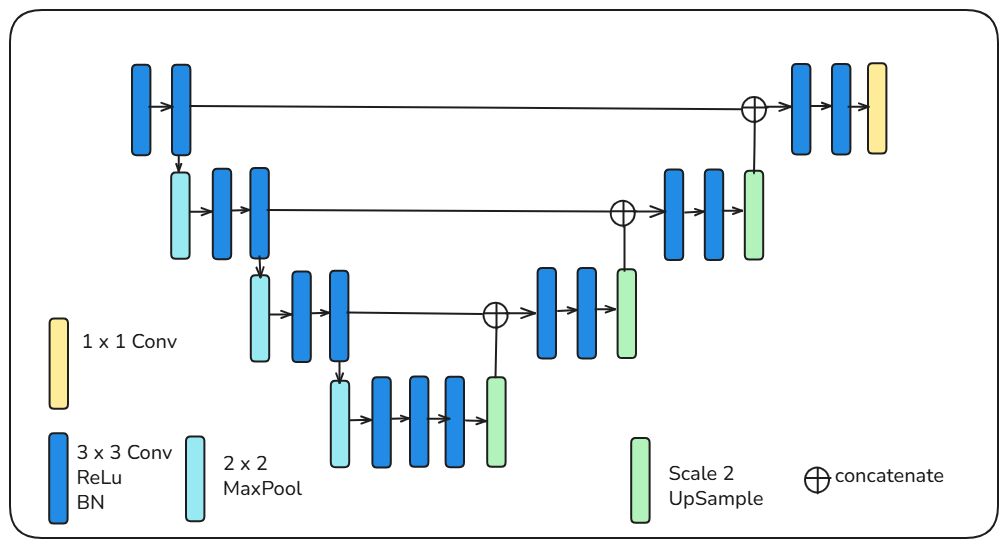} 
  \caption{Detailed architecture of the V2 Encoder-Decoder.}
  \label{fig:v2_structure} 
\end{figure}

\textbf{Motion-Guided Feature Encoding.} Given an input sequence of three consecutive frames $\mathcal{I} \in \mathbb{R}^{3 \times 3 \times H \times W}$, the processing initiates with the MDD module. As detailed in Section 3.1, MDD extracts directional polarity fields, which are immediately interleaved with the raw RGB frames to form a motion-augmented tensor $\mathcal{X}_{in}$. This composite input is fed into the V2 Backbone (detailed in Figure \ref{fig:v2_structure}), which employs a U-Net\cite{2015UNet}-like topology to capture multi-scale spatio-temporal contexts. The encoder progressively downsamples the feature space to a bottleneck representation, while the decoder reconstructs the spatial details via skip connections.

\textbf{Residual-Based Refinement.} Distinct from conventional architectures that treat the decoder's output as the final prediction, TrackNetV5 treats it as a \textit{preliminary coarse draft} containing latent structural information. To recover fine-grained trajectory details, we introduce a long-range skip connection that bridges the front-end MDD module directly to the back-end R-STR module. This design allows the refinement head to leverage both the decoder's draft predictions and the original high-frequency motion cues. The R-STR module aggregates these features to predict a residual correction map, yielding the final high-precision heatmaps.

\subsection{Motion Direction Decoupling Module}

\begin{figure}[t]
  \centering
  \includegraphics[width=0.9\linewidth]{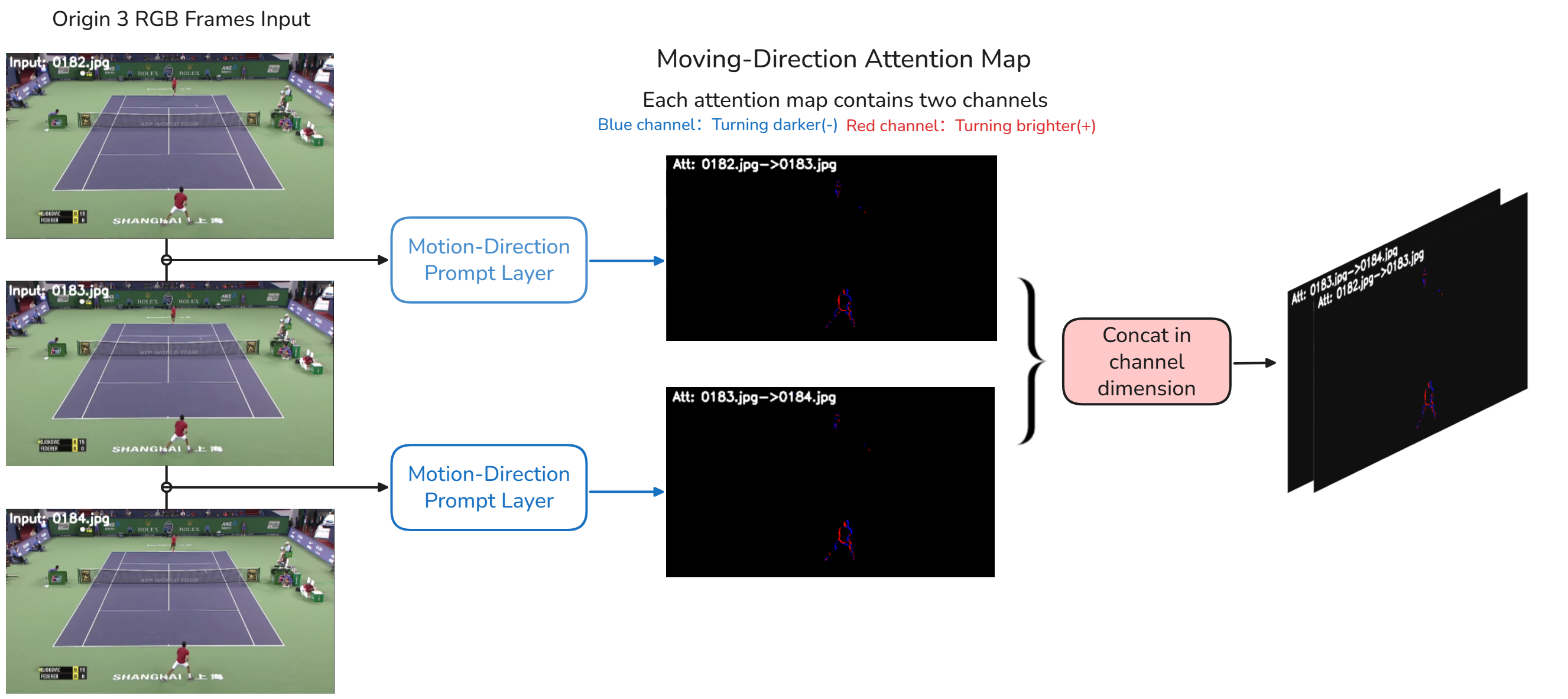}
  \caption{The overall architecture of the Motion Direction Decoupling (MDD) mechanism. }
  \label{fig:mdd_architecture}
\end{figure}

Standard frame-differencing approaches efficiently capture motion occurrences but suffer from \textit{directional ambiguity}, as the absolute difference operation inherently discards the polarity of intensity changes. To resolve this limitation, we propose the Motion Direction Decoupling (MDD) module. As illustrated in Figure \ref{fig:mdd_architecture}, this module decomposes temporal dynamics into distinct polarity fields to recover directional priors, which are then fused with visual features to guide the backbone network.

\subsubsection{Directional Polarity Decomposition}
Consider an input sequence of three consecutive frames $I_{t-1}, I_t, I_{t+1}$. The temporal evolution between adjacent frames is first captured via raw difference maps $D_{t-1 \rightarrow t} = I_t - I_{t-1}$ and $D_{t \rightarrow t+1} = I_{t+1} - I_t$. 

\begin{figure}[t]
  \centering
  \includegraphics[width=0.9\linewidth]{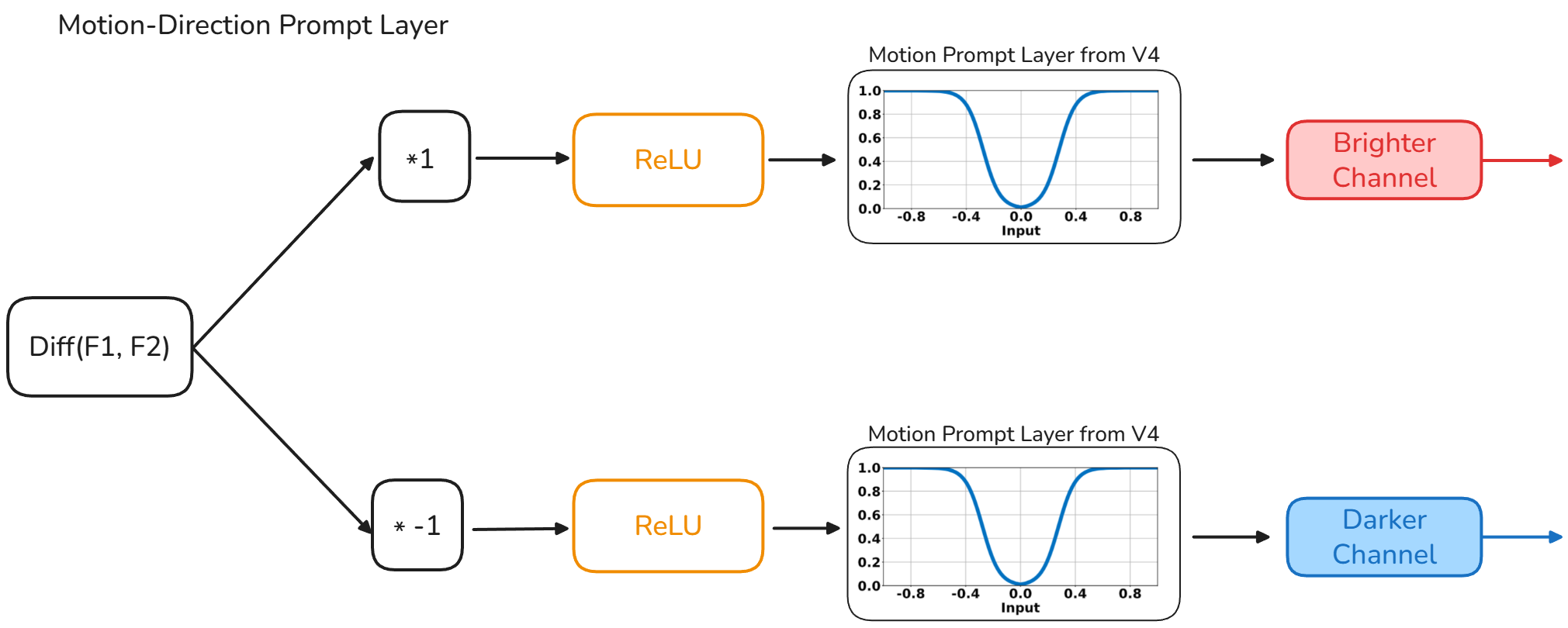}
  \caption{Schematic illustration of the Polarity Decomposition Layer.}
  \label{fig:polarity_logic}
\end{figure}

Unlike previous methods that utilize the magnitude $|D|$, we disentangle the motion signals into positive (brightening) and negative (darkening) polarity channels. As shown in Figure \ref{fig:polarity_logic}, this operation is formulated as:

\begin{equation}
    P^{+}(\Delta) = \text{ReLU}(\Delta), \quad P^{-}(\Delta) = \text{ReLU}(-\Delta)
\end{equation}

where $\Delta$ represents the raw difference map. By applying this decomposition, we obtain paired polarity maps for each temporal interval, effectively encoding the trajectory direction—specifically, a moving object creates a negative trace at its departure point ($P^-$) and a positive trace at its arrival point ($P^+$).

\subsubsection{Learnable Attention Mapping}
The raw pixel intensities in polarity maps $P^{+}$ and $P^{-}$ vary significantly with lighting conditions. To transform these signals into robust attention weights constrained to $[0, 1]$, we employ a parameterized non-linear mapping function $f(\cdot)$. For a given polarity intensity $x$, the attention weight $A$ is computed as:

\begin{equation}
    A = f(x; \alpha, \beta) = \frac{1}{1 + \exp\left(-k(\alpha) \cdot (|x| - m(\beta))\right)}
\end{equation}

Here, $k(\alpha)$ and $m(\beta)$ control the slope and center offset of the activation function, respectively. They are derived from learnable parameters $\alpha$ and $\beta$ to ensure numerical stability and adaptivity:
\begin{equation}
    k(\alpha) = \frac{5.0}{0.45 |\tanh(\alpha)| + \epsilon}, \quad m(\beta) = 0.6 \tanh(\beta)
\end{equation}
where $\epsilon$ is a small constant. This design allows the network to adaptively determine the optimal sensitivity threshold for motion activation during backpropagation, filtering out background noise while highlighting fast-moving objects.

\subsubsection{Motion-Aware Feature Aggregation}
The generated attention maps are explicitly integrated into the network's input stream to guide feature extraction. Let $\mathcal{A}_{t-1, t} \in \mathbb{R}^{2 \times H \times W}$ denote the concatenated attention maps $\{A(P^{+}_{t-1 \rightarrow t}), A(P^{-}_{t-1 \rightarrow t})\}$ for the first interval. 

To preserve both spatial appearance and temporal dynamics, the input to the backbone encoder is constructed by interleaving raw RGB frames with their corresponding motion attention maps:

\begin{equation}
    \mathcal{X}_{in} = \text{Concat}\left( I_{t-1}, \mathcal{A}_{t-1, t}, I_t, \mathcal{A}_{t, t+1}, I_{t+1} \right)
\end{equation}

This results in a 13-channel tensor (assuming 3-channel RGB frames), ensuring that the subsequent convolutional layers possess aligned spatial-temporal contexts. Furthermore, as shown in our architecture, these directional attention maps are also reused in the prediction head to spatially gate the final output, enhancing the model's focus on valid motion regions.

\subsection{Residual-Driven Spatio-Temporal Refinement}

Existing approaches in the TrackNet family predominantly rely on frame-independent predictions, largely neglecting the rich temporal consistency inherent in ball trajectories. To address this limitation, we propose the Residual-Driven Spatio-Temporal Refinement (R-STR) module. The core insight is that the preliminary heatmap drafts generated by the MIMO architecture contain latent spatio-temporal contexts that can be leveraged for two distinct refinement tasks: \textit{temporal enhancement} for low-confidence occlusion frames and \textit{noise suppression} for motion artifacts.

\subsubsection{R-STR Structure}

\begin{figure}[t] \centering \includegraphics[width=0.9\linewidth]{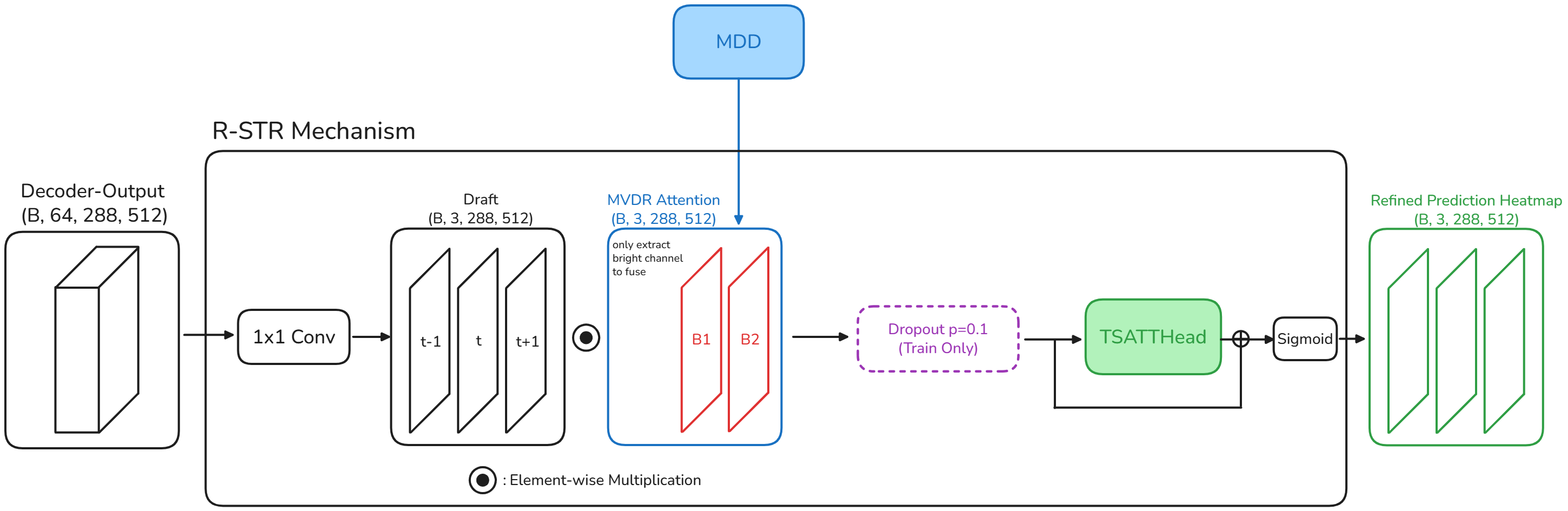} \caption{The workflow of the Residual-Driven Spatio-Temporal Refinement (R-STR) module. The Dropout mechanism is active only during the training phase.} \label{fig:figX} \end{figure}

To effectively integrate these spatio-temporal cues, the R-STR module operates on a residual learning paradigm. As illustrated in Figure \ref{fig:figX}, the workflow proceeds in three stages. First, the decoder features are compressed via a 1$\times$1 convolution to generate preliminary heatmap drafts. Second, to incorporate motion priors, these drafts are fused with the motion attention maps from the MDD module, yielding a motion-aware feature map denoted as $Draft_{MDD}$.

\begin{figure}[t] \centering \includegraphics[width=0.9\linewidth]{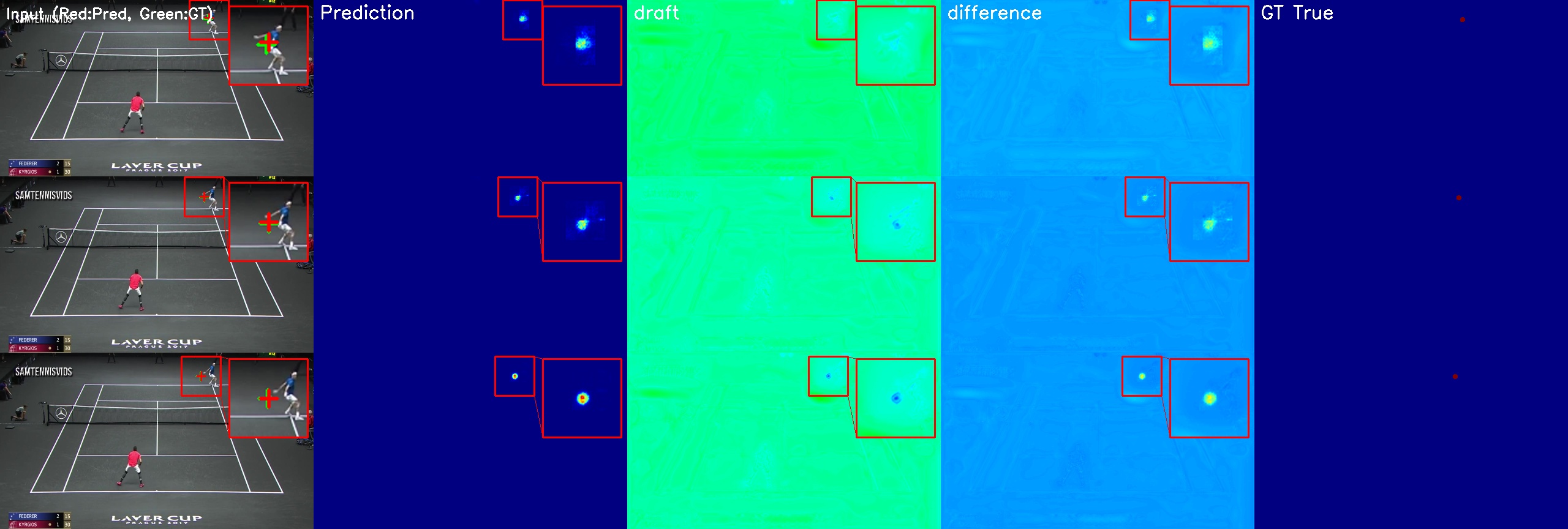} \caption{The visualization of the Residual-Driven Spatio-Temporal Refinement (R-STR) module.} \label{fig:figV} \end{figure}

Finally, instead of reconstructing the final probability map from scratch, we employ a dedicated Spatio-Temporal Self-Attention Head (TSATTHead) to estimate a correction tensor $\Delta$. We posit that learning a corrective residual is significantly more convergent than full reconstruction.

To enforce robust temporal dependency learning, we introduce the \textbf{Stochastic Context Masking} strategy, realized by applying a dropout layer ($\rho=0.1$) to the $Draft_{MDD}$ \textbf{only during training} (as shown in Figure \ref{fig:figX}). We define the masked input as $Draft'_{MDD} = \text{Dropout}(Draft_{MDD})$.

The final heatmap refinement is formulated according to the stage:

\textbf{During Training}, the Dropout mechanism is active, and the final prediction is calculated from the corrupted draft $Draft'_{MDD}$. This forces the TSATTHead to learn a strong reconstruction residual $\Delta$ to recover the randomly zeroed-out pixels.
$$ H_{final}^{\text{train}} = \sigma(\mathbf{Draft'_{MDD}} + \Delta^{\text{train}}) $$
The residual tensor $\Delta^{\text{train}}$ is estimated from the masked input:
$$ \Delta^{\text{train}} = \text{TSATTHead}(\mathbf{Draft'_{MDD}}) $$

\textbf{During Inference}, the Dropout is disabled ($\rho=0$), and the model operates on the clean draft $Draft_{MDD}$. The correction tensor $\Delta^{\text{inference}}$ is applied to the clean base.
$$ H_{final}^{\text{inference}} = \sigma(\mathbf{Draft_{MDD}} + \Delta^{\text{inference}}) $$
The residual tensor $\Delta^{\text{inference}}$ is estimated from the clean input:
$$ \Delta^{\text{inference}} = \text{TSATTHead}(\mathbf{Draft_{MDD}}) $$

where $\sigma$ denotes the Sigmoid activation function. This adaptive residual design ensures that the network is maximally regularized during training for robustness while maintaining optimal prediction quality on the clean input during deployment.

\subsubsection{TSATTHead Architecture}

\begin{figure}[h]
  \centering
  \includegraphics[width=0.9\linewidth]{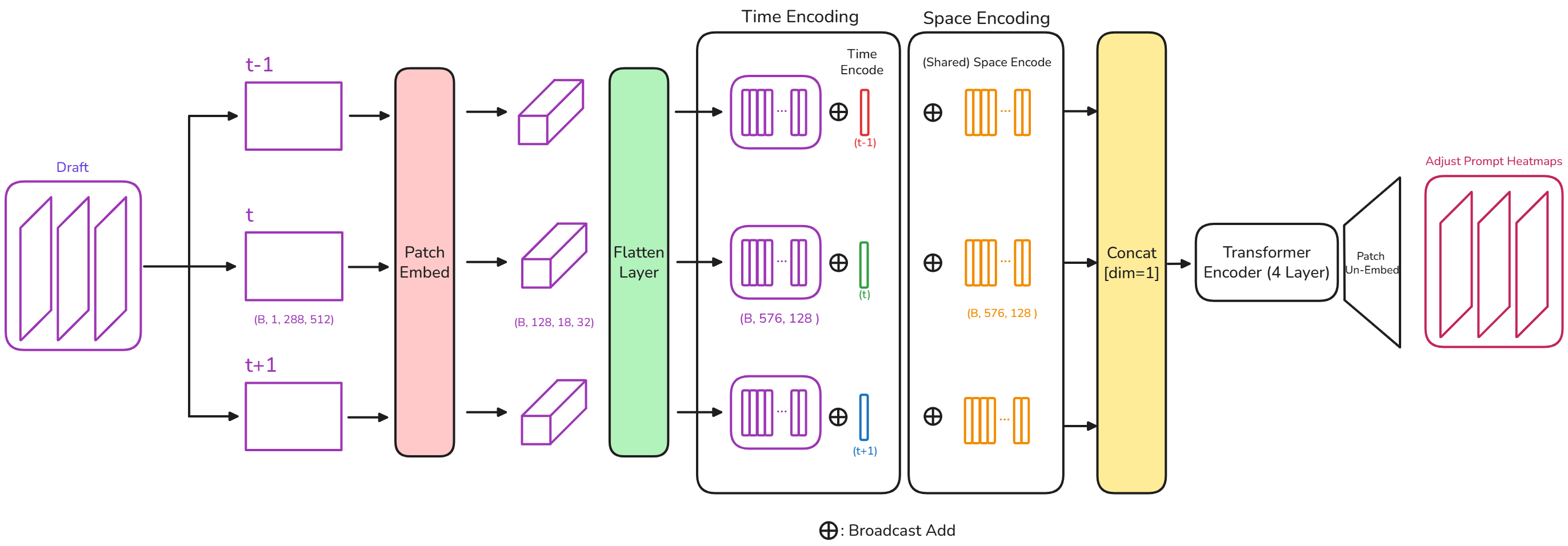}
  \caption{TSATTHead Design.}
  \label{fig:figY}
\end{figure}

The TSATTHead (refer to Figure \ref{fig:figY}) serves as the computational engine for estimating the residual $\Delta$. It is designed as a lightweight Transformer that refines the initial drafts through a serialized pipeline.


Following the masking stage, we employ a patch embedding strategy \cite{2021ViT} where non-overlapping patches are flattened and projected into an embedding space. These tokens are enriched with factorized spatio-temporal positional encodings to preserve sequence order. The sequence is then processed by a Transformer Encoder \cite{2017Attention}, where multi-head self-attention enables global interaction across spatial and temporal dimensions. Specifically, inspired by the factorized attention mechanism of TimeSformer \cite{2021TimeSformer}, we decompose the self-attention into separate spatial and temporal blocks to efficiently capture and reconstruct the trajectory coherence. Finally, the processed tokens are decoded via a PixelShuffle \cite{2016ESPCN} operation to generate the high-resolution residual map $\Delta$, which is superposed onto the original clean draft to yield the final prediction.

\section{Experiment}

\subsection{Implementation Details}

\textbf{Datasets and Inputs.}
We evaluate our method on two datasets: the public TrackNetV2 Dataset ($1280 \times 720$) and our internal Loveall Dataset ($1920 \times 1080$). Following standard protocols \cite{2020TrackNetV2}, the data is split into training and validation sets with a 7:3 ratio. During training and inference, all input frames are resized to a resolution of $512 \times 288$ to balance efficiency and spatial precision. The model takes a sequence of three consecutive RGB frames ($I_{t-1}, I_t, I_{t+1}$) as input to predict the trajectory heatmaps.

\textbf{Ground Truth Generation.}
Unlike methods using soft Gaussian distributions, we employ a Gaussian-guided binary masking strategy to sharpen supervision signals. For a ground truth ball coordinate $(c_x, c_y)$, we first define a region of interest (ROI) using a Gaussian kernel radius $r$. To mitigate class imbalance, we binarize this region:
\begin{equation}
    Y(x, y) = 
    \begin{cases} 
    1, & \text{if } (x, y) \in \text{ROI}_{r} \\
    0, & \text{otherwise}
    \end{cases}
\end{equation}
We set the kernel radius $r=30$ for the TrackNetV2 dataset and $r=40$ for the higher-resolution Loveall dataset, ensuring consistent supervision scales relative to the ball size.

\textbf{Loss Function.}
Following TrackNetV2 \cite{2020TrackNetV2}, we utilize the Weighted Binary Cross Entropy (WBCE). This loss function addresses the extreme foreground-background imbalance by dynamically weighting the pixel-wise entropy based on prediction confidence. The formulation is defined as:
\begin{equation}
    L_{WBCE} = - \frac{1}{N} \sum_{i} \left[ (1 - p_i)^2 y_i \log(p_i) + p_i^2 (1 - y_i) \log(1 - p_i) \right]
\end{equation}
where $p_i$ represents the predicted probability and $y_i \in \{0, 1\}$ is the ground truth label. The weights $(1-p_i)^2$ and $p_i^2$ enforce the model to focus on hard examples, preventing the vast background regions from dominating the gradient.

\textbf{Training Setup.}
The model is implemented in PyTorch and trained on a single NVIDIA RTX 4090 (24GB) GPU. We use the AdamW optimizer\cite{2019AdamW} with a batch size of 2 and an initial learning rate of $1 \times 10^{-4}$. The training spans 30 epochs, with a multi-step learning rate decay ($\gamma=0.1$) applied at epochs 20 and 25. 

\textbf{Evaluation Metrics.}
For fair benchmarking, all inference speed tests (FPS) reported in this paper are conducted on an NVIDIA T4 GPU. Following the standard TrackNet evaluation protocol, a prediction is defined as a True Positive (TP) if the Euclidean distance between the predicted center and the ground truth center is within a tolerance of \textbf{4 pixels} at the original resolution. Precision, Recall, and F1-score are calculated based on this criterion.

\subsection{Main Results}

We evaluate TrackNetV5 against state-of-the-art methods (TrackNetV2 and TrackNetV4) on the standard TrackNetV2 dataset and the Loveall dataset. Results are summarized in Table \ref{tab:results_tracknetv2}, Table \ref{tab:results_loveall}, and Table \ref{tab:performance}.

\textbf{Comparison on Public Benchmarks.} 
As shown in Table \ref{tab:results_tracknetv2}, TrackNetV5 achieves a new state-of-the-art F1-score of \textbf{0.9859} and an Accuracy of \textbf{0.9733}, outperforming the previous best TrackNetV4 by 2.78\% in F1. A critical analysis of the Precision-Recall trade-off highlights our architectural advantage: while V4 attains high Precision (0.9965), it suffers from conservative predictions with low Recall (0.9225) and 1,317 False Negatives. In contrast, TrackNetV5 drastically reduces False Negatives by 73.9\% (to 344) while maintaining high Precision (0.9923). This confirms that the MDD and R-STR modules effectively recover hard-to-detect targets without introducing excessive noise.
Furthermore, on the Loveall Dataset (Table \ref{tab:results_loveall}), TrackNetV5 demonstrates superior generalization with an F1-score of \textbf{0.9878}. The reduction of False Negatives to just 186 (vs. 523 for V4) proves that our explicit directional modeling and temporal refinement are robust to domain shifts in lighting and camera angles.

\begin{table}[t]
    \caption{Quantitative comparison on the \textbf{TrackNetV2 Dataset}. The best results are highlighted in \textbf{bold}.}
    \centering
    \resizebox{\linewidth}{!}{
    \begin{tabular}{lccccccccccc}
        \toprule
        Model & \textbf{Acc} & \textbf{Precision} & \textbf{Recall} & \textbf{F1} & Total & TP & FP1 & FP2 & FP & TN & FN \\
        \midrule
        V2 & 0.9396 & 0.9919 & 0.9446 & 0.9677 & 17682 & 15988 & 108 & 23 & 131 & 626 & 937 \\
        V4 & 0.9224 & \textbf{0.9965} & 0.9225 & 0.9581 & 17682 & 15669 & 47 & 8 & 55 & 641 & 1317 \\
        V5 (Ours) & \textbf{0.9733} & 0.9923 & \textbf{0.9797} & \textbf{0.9859} & 17682 & 16573 & 116 & 13 & 129 & 636 & 344 \\
        \bottomrule
    \end{tabular}
    }
    \label{tab:results_tracknetv2}
\end{table}

\begin{table}[t]
    \caption{Generalization test on the \textbf{Loveall Dataset}.}
    \centering
    \resizebox{\linewidth}{!}{
    \begin{tabular}{lccccccccccc}
        \toprule
        Model & \textbf{Acc} & \textbf{Precision} & \textbf{Recall} & \textbf{F1} & Total & TP & FP1 & FP2 & FP & TN & FN \\
        \midrule
        V2 & 0.9542 & \textbf{0.9947} & 0.9584 & 0.9762 & 11313 & 10615 & 54 & 3 & 57 & 180 & 461 \\
        V4 & 0.9484 & 0.9943 & 0.9528 & 0.9731 & 11313 & 10555 & 52 & 9 & 61 & 174 & 523 \\
        V5 (Ours) & \textbf{0.9763} & 0.9925 & \textbf{0.9832} & \textbf{0.9878} & 11313 & 10864 & 80 & 2 & 82 & 181 & 186 \\
        \bottomrule
    \end{tabular}
    }
    \label{tab:results_loveall}
\end{table}

\textbf{Efficiency vs. Accuracy Trade-off.} 
Finally, we analyze the computational cost on an NVIDIA T4 GPU (Table \ref{tab:performance}). While TrackNetV5 introduces additional modules, the overhead is marginal, with only a 3.7\% increase in FLOPs (117.09 G) compared to V4. Crucially, it maintains a real-time inference speed of 114 FPS ($38.12 \times 3$), far exceeding the requirement for standard sports broadcasting (30/60 FPS). This demonstrates that TrackNetV5 achieves a highly favorable trade-off, delivering substantial accuracy gains with negligible computational cost.

\begin{table}[h]
    \caption{Efficiency comparison on NVIDIA T4. FPS is measured with real video inference.}
    \centering
    \begin{tabular}{lccc}
        \toprule
        Model & \textbf{FPS (Real)} & \textbf{FLOPs / G} & \textbf{Params / M} \\
        \midrule
        V2 & $41.09 \times 3$ & 112.89 & 11.33 \\
        V4 & $40.32 \times 3$ & 112.89 & 11.33 \\
        V5 (Ours) & $38.12 \times 3$ & 117.09 & 14.77 \\
        \bottomrule
    \end{tabular}
    \label{tab:performance}
\end{table}

\subsection{Ablation Study}

\textbf{Effectiveness of Motion Direction Decoupling (MDD).}
We first verify the impact of the MDD module by comparing the baseline TrackNetV2 with V2 equipped with MDD. As shown in Table \ref{tab:ablation_mdd}, the MDD module improves the F1-score from 0.9677 to 0.9695 and reduces False Negatives from 937 to 861. This indicates that explicitly modeling directional polarity helps the network recall ambiguous targets that are overlooked by the baseline.

\begin{table}[h]
    \caption{Ablation study on the \textbf{MDD Module} using the TrackNet dataset. MDD improves recall and F1-score compared to the V2 and V4 baselines.}
    \centering
    \resizebox{\linewidth}{!}{
    \begin{tabular}{lccccccccccc}
        \toprule
        Model & \textbf{Acc} & \textbf{Precision} & \textbf{Recall} & \textbf{F1} & Total & TP & FP1 & FP2 & FP & TN & FN \\
        \midrule
        V2 & 0.9396 & 0.9919 & 0.9446 & 0.9677 & 17682 & 15988 & 108 & 23 & 131 & 626 & 937 \\
        V4 & 0.9224 & \textbf{0.9965} & 0.9225 & 0.9581 & 17682 & 15669 & 47 & 8 & 55 & 641 & 1317 \\
        V2 + MDD & \textbf{0.9428} & 0.9907 & \textbf{0.9491} & \textbf{0.9695} & 17682 & \textbf{16047} & 125 & 25 & 150 & 624 & 861 \\
        \bottomrule
    \end{tabular}
    }
    \label{tab:ablation_mdd}
\end{table}

\textbf{Effectiveness of R-STR Module and Final Architecture Choice.}
We further evaluate the R-STR module. As presented in Table \ref{tab:ablation_rstr}, the "V2 + R-STR" configuration achieves a remarkable performance boost, with the F1-score reaching 0.9866. 

\begin{table}[h]
    \caption{Ablation study on the \textbf{R-STR Module}. While "V2 + R-STR" achieves the highest F1, it incurs higher False Positives compared to the final TrackNetV5.}
    \centering
    \resizebox{\linewidth}{!}{
    \begin{tabular}{lccccccccccc}
        \toprule
        Model & \textbf{Acc} & \textbf{Precision} & \textbf{Recall} & \textbf{F1} & Total & TP & FP1 & FP2 & FP & TN & FN \\
        \midrule
        V2 & 0.9396 & \textbf{0.9919} & 0.9446 & 0.9677 & 17682 & 15988 & 108 & 23 & 131 & 626 & 937 \\
        V2 + R-STR & \textbf{0.9745} & 0.9885 & \textbf{0.9848} & \textbf{0.9866} & 17682 & 16607 & 169 & 25 & 194 & 624 & 257 \\
        \bottomrule
    \end{tabular}
    }
    \label{tab:ablation_rstr}
\end{table}

It is worth noting that the "V2 + R-STR" configuration yields a marginally higher F1-score (0.9866) than our final TrackNetV5 (0.9859, which combines MDD and R-STR). However, we deliberately selected the combined architecture as the final TrackNetV5 for practical considerations.
As shown in comparisons between Table \ref{tab:results_tracknetv2} and Table \ref{tab:ablation_rstr}, although "V2 + R-STR" has excellent recall, its Precision is approximately 0.39\% lower than the final V5 (0.9885 vs. 0.9923), and it generates 65 more False Positives (194 vs. 129). In real-world sports analytics, high Precision is often prioritized to minimize false alarms, which can disrupt automated game analysis. By integrating MDD, the final TrackNetV5 effectively suppresses these outliers, offering a more robust and commercially viable balance between Precision and Recall.

\section{Conclusion}

In this paper, we addressed a long-standing challenge in sports video analytics: high-speed small object tracking. Motivated by two major deficiencies in the existing TrackNet family (a lack of explicit motion directionality representation and the failure to utilize wasted spatio-temporal context in the MIMO architecture), we proposed TrackNetV5, a new model introducing two complementary architectural innovations. First, we designed the MDD module, which successfully encodes movement direction information into the motion features by decoupling change polarity. Second, we introduced the R-STR module, which leverages a lightweight spatio-temporal Transformer and a residual error-correction mechanism to, for the first time, exploit the spatio-temporal context among multi-frame heatmap drafts.

Extensive experiments on the public TrackNet tennis dataset and our proprietary LoveAll dataset (which features low-angle baseline views, more occlusions, and greater difficulty) demonstrate the effectiveness of our method. TrackNetV5 significantly outperforms all prior benchmarks, establishing a new SOTA. More importantly, this performance leap is achieved with minimal computational overhead (GFLOPs increased by only \textbf{3.7\%} over V2 and V4), highlighting the high efficiency of our design.

Although our work has limitations and potential for improvement—such as investigating whether a larger temporal window could further suppress False Positives in complex backgrounds—we believe the efficient motion representation and spatio-temporal refinement paradigms demonstrated by TrackNetV5 offer valuable insights for real-time small object tracking and other complex sports analytics tasks.

\section*{Additional Visualization}

\begin{figure}[htb!]
    \centering
    \includegraphics[width=1.0\linewidth]{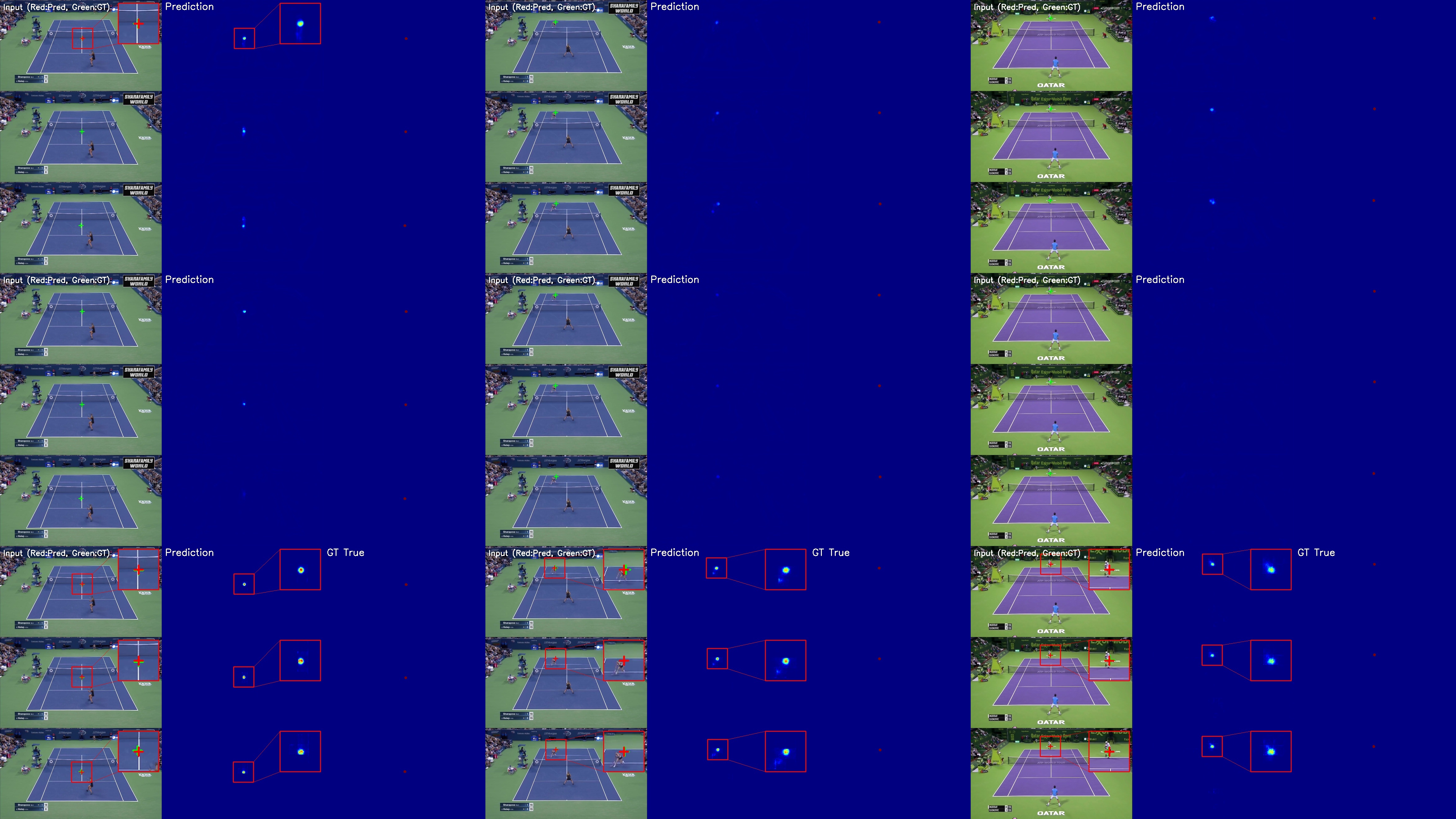}
    \caption{Comparative visualization of models V2, V4, and V5 on TrackNet Dataset. Each block of three rows corresponds to one model.}
    \label{fig:figexample2}
\end{figure}

\begin{figure}[htb!]
    \centering
    \includegraphics[width=1.0\linewidth]{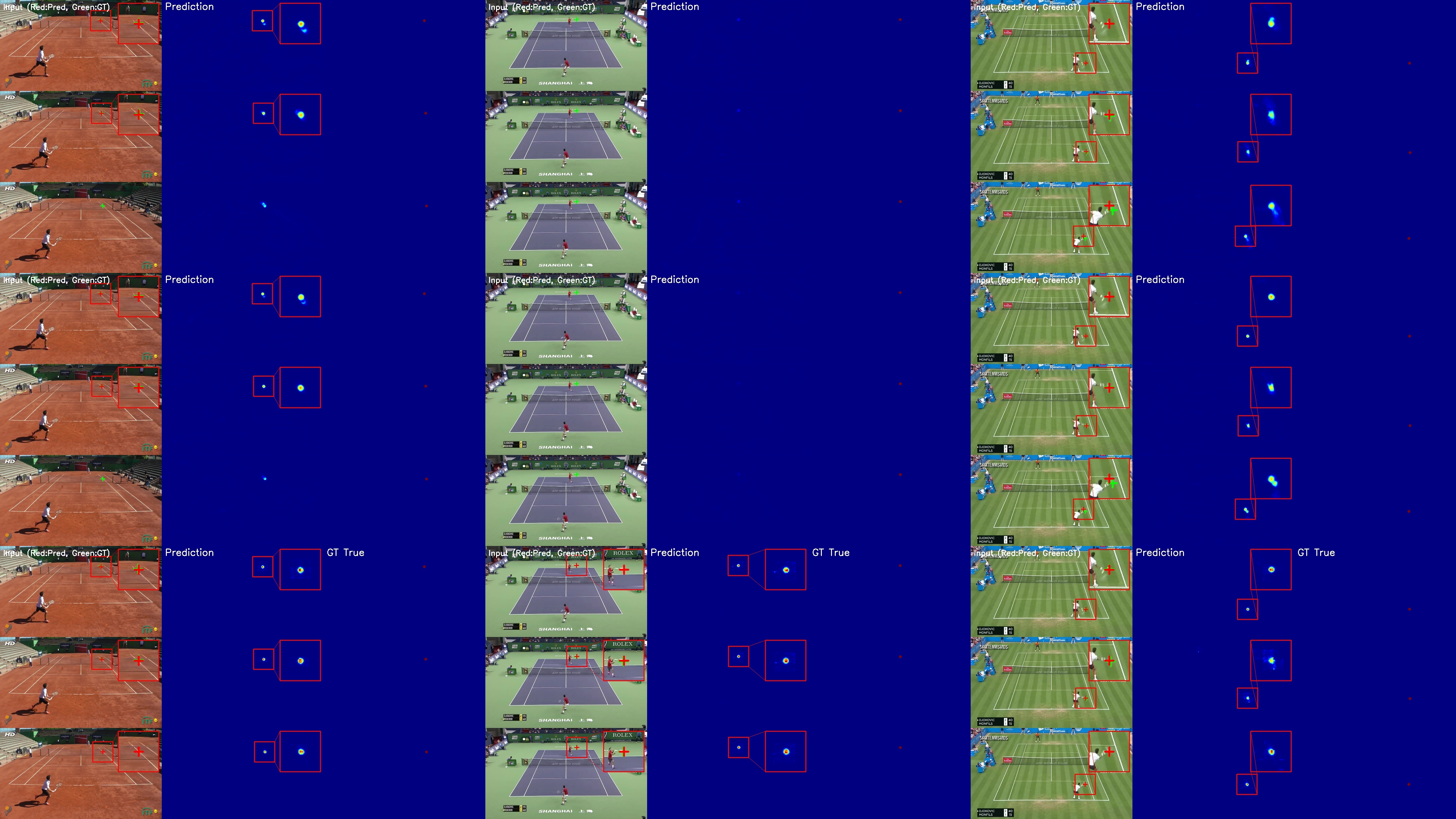}
    \caption{Comparative visualization of models V2, V4, and V5 on TrackNet Dataset. Each block of three rows corresponds to one model.}
    \label{fig:figexample4}
\end{figure}

\begin{figure}[htb!]
    \centering
    \includegraphics[width=1.0\linewidth]{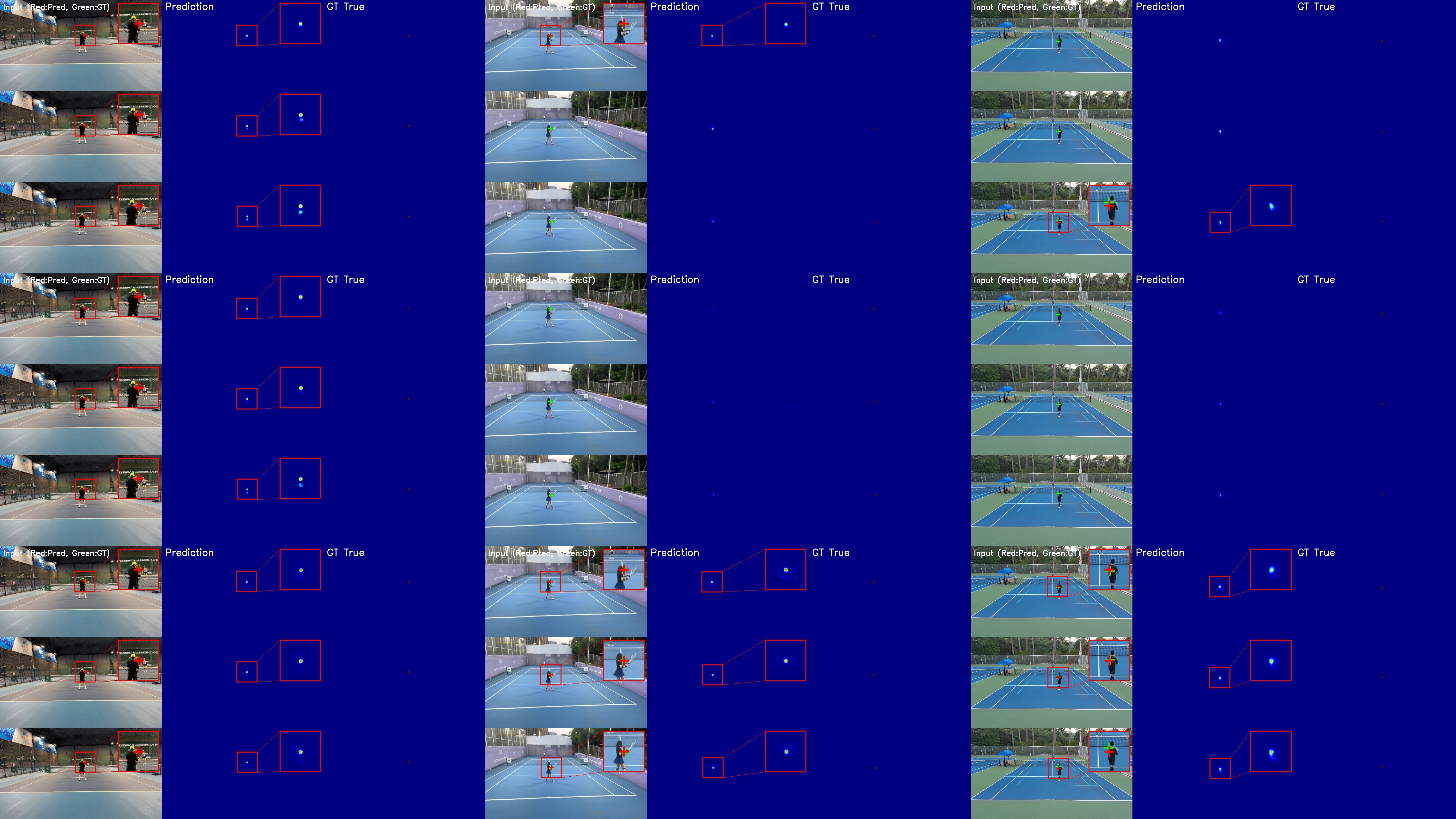}
    \caption{Comparative visualization of models V2, V4, and V5 on LoveAll Dataset. Each block of three rows corresponds to one model.}
    \label{fig:loveallexample1}
\end{figure}

\begin{figure}[htb!]
    \centering
    \includegraphics[width=1.0\linewidth]{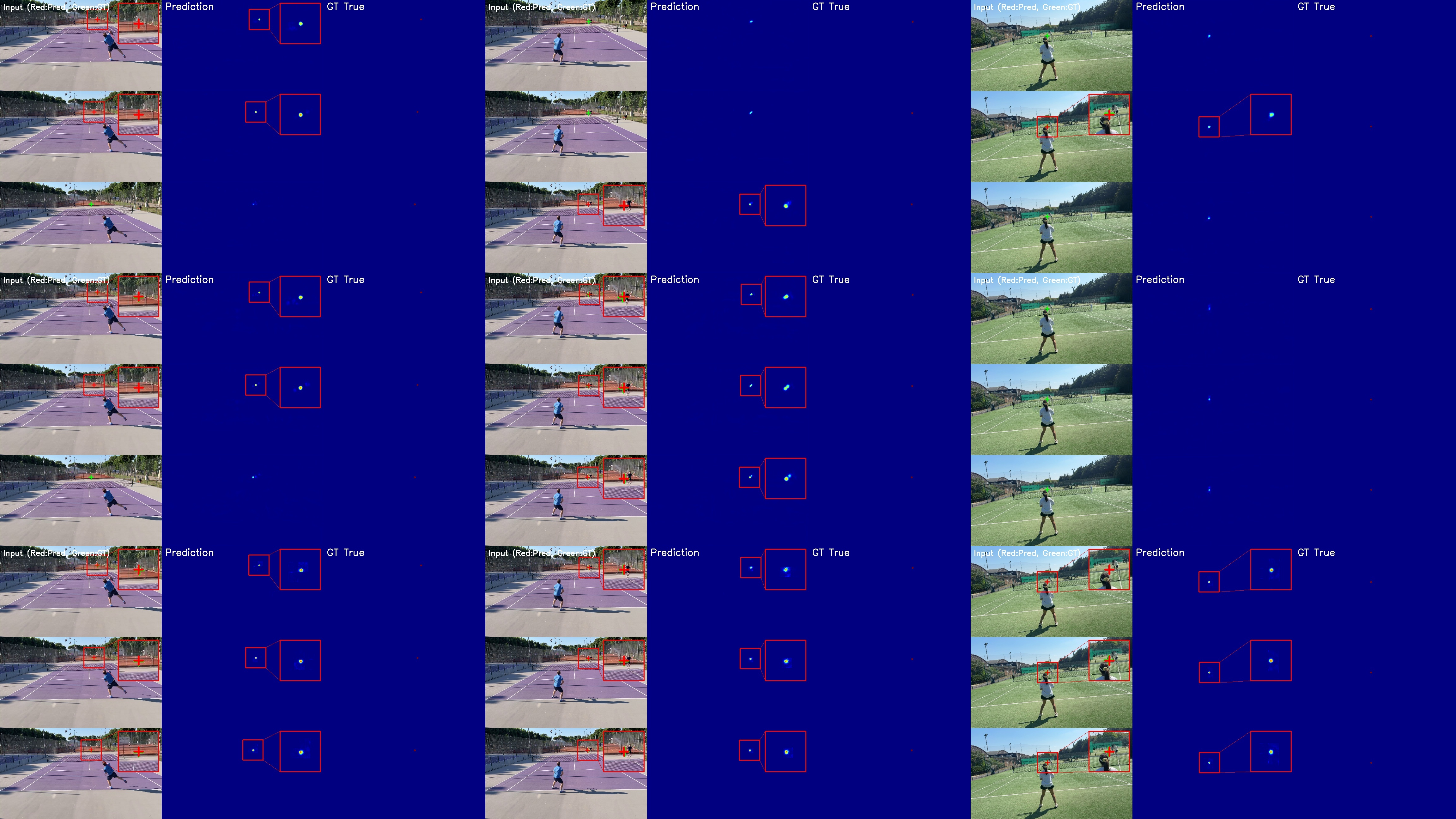}
    \caption{Comparative visualization of models V2, V4, and V5 on LoveAll Dataset. Each block of three rows corresponds to one model.}
    \label{fig:loveallexample2}
\end{figure}

The effectiveness of models V2, V4, and V5 in tennis ball tracking is further substantiated by the comparative results presented in the figures below. These figures, including Figure \ref{fig:figexample1}, Figure \ref{fig:figexample2}, and Figure \ref{fig:figexample4} from TrackNet Dataset, and Figure \ref{fig:loveallexample1} and Figure \ref{fig:loveallexample2} from LoveAll Dataset serve as qualitative examples of the models' performance.

Each visualization panel within a figure is structured into three consecutive vertical sections to facilitate direct comparison:
\begin{itemize}
    \item The \textbf{top three rows} illustrate the results produced by the \textbf{V2} model.
    \item The \textbf{middle three rows} display the predictions from the \textbf{V4} model.
    \item The \textbf{bottom three rows} correspond to the output of the \textbf{V5} model.
\end{itemize}

The prediction accuracy is visually assessed through distinct markers superimposed on the frames:
\begin{itemize}
    \item A \textbf{green cross} indicates the \textbf{Ground Truth (GT)} position of the tennis ball.
    \item A \textbf{red cross} denotes the position predicted by the respective model (\textbf{Prediction}).
\end{itemize}

Crucially, in instances where the green cross is not visible, it signifies that the model's red prediction cross has achieved exact spatial overlap with the true ball position. This complete superimposition visually confirms a zero-error prediction for that specific frame, highlighting the precision attained by the model.


\begin{figure}[htb!] 
    \centering
    \includegraphics[width=1.0\linewidth]{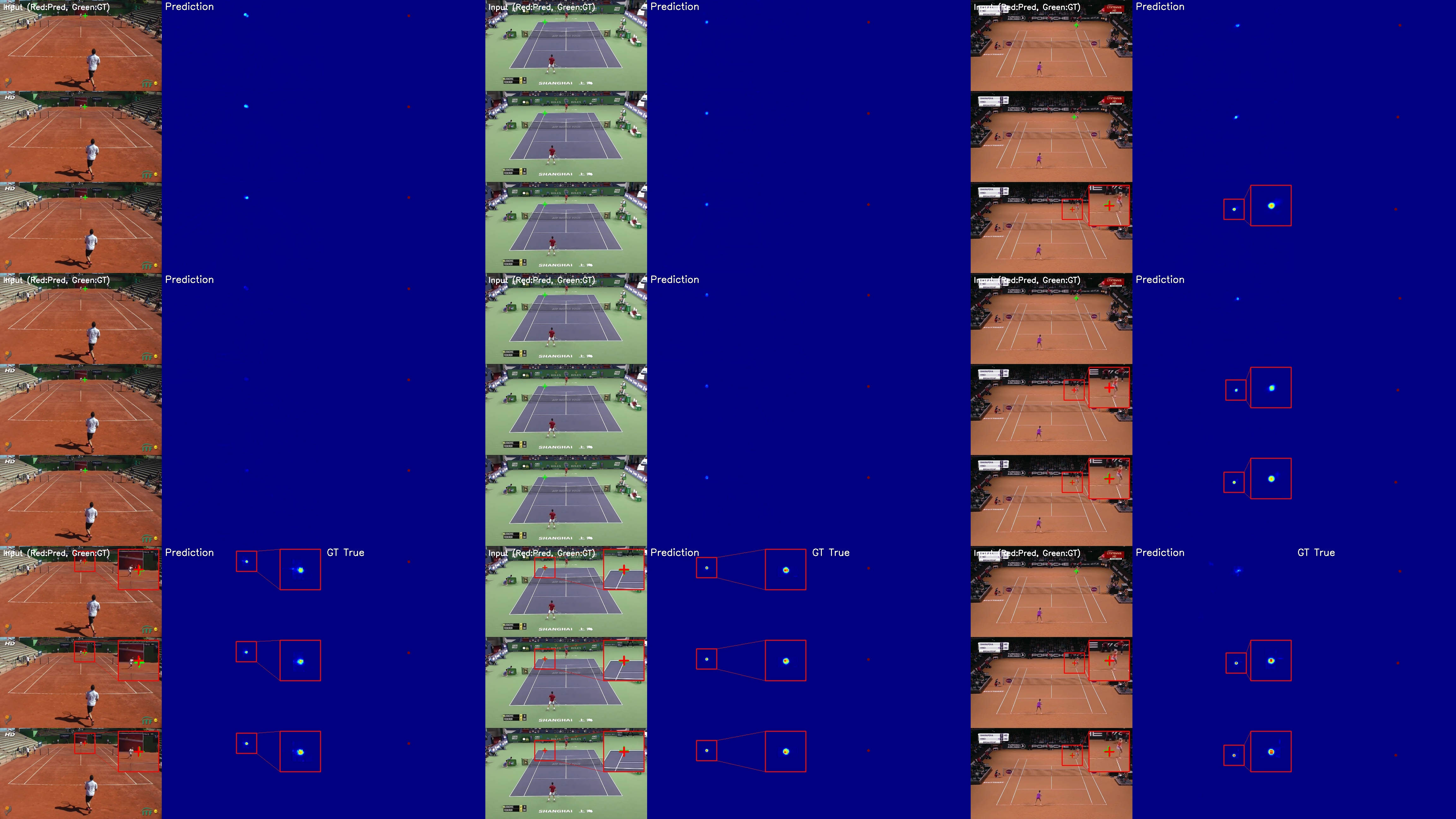}
    \caption{Comparative visualization of models V2, V4, and V5 on TrackNet Dataset. Each block of three rows corresponds to one model.}
    \label{fig:figexample1}
\end{figure}

\clearpage

\section*{Acknowledgments}
This work was supported by Shanghai Code Zero Sports Technology Co., which provided funding, computational resources, internal datasets, and technical guidance for this project. 

\bibliographystyle{unsrt}  
\bibliography{references}  

\end{document}